\documentclass[sigconf,natbib=true,anonymous=false]{acmart}
\usepackage[inline]{enumitem}
\usepackage{enumitem}

\AtBeginDocument{%
  }

\setcopyright{acmlicensed}
\copyrightyear{2024}
\acmYear{2024}
\acmDOI{XXXXXXX.XXXXXXX}

\acmConference[SIGIR '24]{the 47th International ACM SIGIR Conference on Research and Development in Information Retrieval}{July 14--18,
  2024}{Washington D.C., USA}
\acmISBN{978-1-4503-XXXX-X/18/06}

\usepackage[linesnumbered, ruled]{algorithm2e}
\usepackage{colortbl}
\usepackage{multirow}
\usepackage{CJKutf8}
\usepackage{tabularx}
\usepackage{pifont}
\usepackage{makecell}
\usepackage{url}

\begin{document}

\title{Boosting Conversational Question Answering with Fine-Grained Retrieval-Augmentation and Self-Check}



\settopmatter{authorsperrow=1} 
\newcommand{\tsc}[1]{\textsuperscript{#1}} 
\author{Linhao Ye, Zhikai Lei, Jianghao	Yin, Qin Chen$^{*}$, Jie Zhou, Liang He
}
\thanks{*Corresponding author, qchen@cs.ecnu.edu.cn.}
\affiliation{
  \institution{School of Computer Science and Technology, East China Normal University}
  \city{Shanghai}
  \country{China}
}
\email{{lhye,kausal,jhyin}@stu.ecnu.edu.cn,{qchen,jzhou,lhe}@cs.ecnu.edu.cn}
\begin{abstract}

Retrieval-Augmented Generation (RAG) aims to generate more reliable and accurate responses, by augmenting large language models (LLMs) with the external vast and dynamic knowledge. Most previous work focuses on using RAG for single-round question answering, while how to adapt RAG to the complex conversational setting wherein the question is interdependent on the preceding context is not well studied. In this paper, we propose a conversation-level RAG (\texttt{ConvRAG}) approach, which incorporates fine-grained retrieval augmentation and self-check for conversational question answering (CQA). In particular, our approach consists of three components, namely conversational question refiner, fine-grained retriever and self-check based response generator, which work collaboratively for question understanding and relevant information acquisition in conversational settings. Extensive experiments demonstrate the great advantages of our approach over the state-of-the-art baselines. Moreover, we also release a Chinese CQA dataset with new features including reformulated question, extracted keyword, retrieved paragraphs and their helpfulness, which facilitates further researches in RAG enhanced CQA.

\end{abstract}


\vspace{-1mm}


\vspace{-1mm}


\maketitle
\vspace{-1mm}
\section{Introduction}
\begin{figure*}[h]
    \centering
    \includegraphics[width=\linewidth]{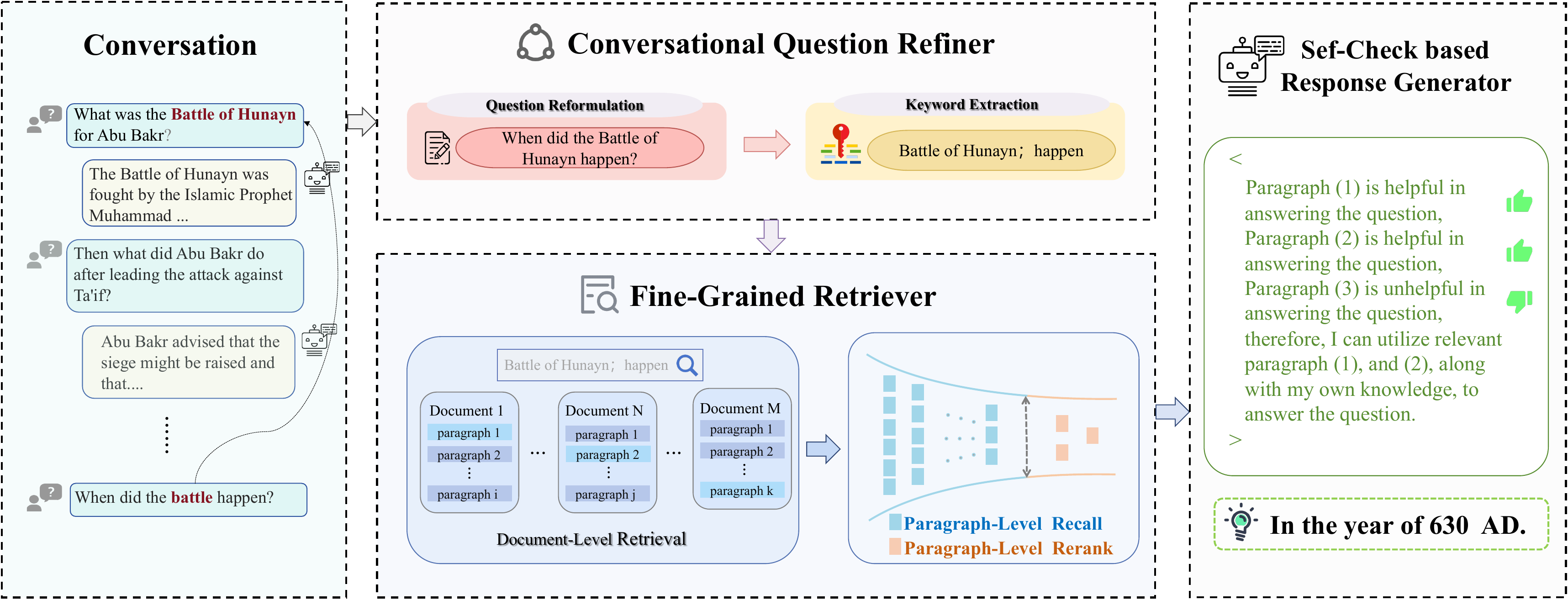}
    \vspace{-7mm}
    \caption{Overall architecture of our \texttt{ConvRAG} approach. }
    \label{figure1}
    \vspace{-2mm}
\end{figure*}

Conversational Question Answering (CQA) has attracted great attention in both academia and industry in recent years, which provides more natural human-computer interactions by extending single-turn question answering (QA) to conversational settings \cite{rajpurkar2016squad, li2023s2m}. In CQA, users usually ask multiple follow-up questions using anaphora that refers to certain concepts in previous conversation history, or ellipsis that can be omitted. As shown in Figure \ref{figure1}, the `battle' in the current question refers to `Hunayn' in the first turn, making it more challenging than single-turn QA.  

One key challenge in CQA is how to explicitly represent the questions based on the interdependent context. Previous work focuses on using the question rewriting methods for a better question understanding. Elgoharyet et al. \cite{elgohary2019can} first released a dataset with human rewrites of questions and analysed the writing quality. Vakulenko et al. \cite{vakulenko2021question} proposed a Transformer-based question rewriting model, and extended evaluation to the impact on the end-to-end CQA task. Recent researches tend to rely on large language models (LLMs) \cite{brown2020language, touvron2023llama, zhang2205opt, anil2023palm} to directly answer the question by modeling the whole context \cite{thoppilan2022lamda}. Whereas, a considerable room for improvement still remains, especially when the topics frequently switch over the conversation according to our preliminary studies.

Another challenge is how to obtain relevant knowledge to answer the open-domain questions. Recently, LLMs have shown great potential in various natural language processing tasks with the encapsulated parametric knowledge. Whereas, they suffer from the hallucination problems and often generate factual errors \cite{bang2023multitask,guerreiro2023hallucinations,huang2023survey,li2023halueval}, which are not applicable for the knowledge-intensive questions. To alleviate this problem, Retrieval-Augmented Generation (RAG) methods are used to augment the input of LLMs with the retrieved information. 
The systems such as WebGPT \cite{nakano2021webgpt}, WebCPM \cite{qin2023webcpm} and WebGLM \cite{liu2023webglm} imitate human behaviors for web search to enhance answer generation. Noting that some useless or off-topic information will be introduced in this process, Asia et al. \cite{asai2023self} proposed a self-reflective RAG framework, which adaptively retrieved passages on-demand for response generation. 
Whereas, these methods mainly focus on the single-turn QA, where retrieval merely involves the current question. Regarding to CQA that questions are interdependent on the conversation history, how to perform effective retrieval to enhance generation has not well been studied.

To resolve the above two challenges, we propose a Conversation-level Retrieval-Augmented Generation (\texttt{ConvRAG}) approach, which boosts CQA with fine-grained retrieval-augmentation and self-check. Specifically, our approach consists of three components, namely Conversational Question Refiner, Fine-Grained Retriever, and Self-Check based Response Generator. We first perform question refinement to better understand the question with the interdependent conversation history. Then, a fine-grained retriever is used to obtain the most relevant information from Webs. In case some noise or irrelevant information still exists in the retrieved results, we present a self-check mechanism, which urges LLMs to rethink the helpfulness of the retrieved information, and selectively utlize the helpful information for more accurate response generation.

We utilize two well-known benchmark datasets for training, and evaluate on both the benchmark test set and our constructed new dataset with diverse topics and types. The experimental results show that we outperform the state-of-the-art RAG enhanced methods and the industry production system in most cases. The main contributions of our work are as follows:
\begin{itemize}[leftmargin=*, align=left]
     \item  We construct a Chinese CQA dataset extending with new features including reformulated question, extracted keyword, retrieved paragraphs and their helpfulness, which helps promote researches in RAG enhanced CQA.
     \item  We explore the potential of integrating LLMs with RAG in the conversation level, and propose \texttt{ConvRAG} to boost CQA with fine-grained retrieval-augmentation and self-check.
    \item  The three components of \texttt{ConvRAG} well resolve the challenges of question representation and relevant knowledge acquisition for CQA, and extensive experiments also show the superiority of our approach over the baselines.
\end{itemize}

\vspace{-2mm}
\section{RELATED WORK}
\textbf{Conversational Question Answering}.
Conversational Question Answering (CQA) is an extension of the single-round QA, which requires to answer based on the conversation history and the current question \cite{huang2020challenges, gekhman2023robustness}. 
As the follow-up questions are probably dependent on preceding context, one research direction focuses on developing co-reference models or query rewriting methods for better question understanding \cite{elgohary2019can, vakulenko2021question, jeffrey2019cast}. 
Other researches aim to utilize more relevant information for CQA.  
 Liu et al. \cite{liu2018knowledge} explored to utilize external knowledge from Wikipedia to enhance conversational agents. Recent approaches tend to augment LLMs with the information retrieved from Webs for response generation, which helps alleviate the hallucination problem \cite{shuster2021retrieval, feng2023trends, ji2023survey}. Whereas, these approaches are subjected to single-round QA, and how to adapt to the conversational settings deserves further researches.
  
 

\textbf{Retrieval-Augmented Generation}.
Retrieval-Augmented Generation (RAG) enhances text generation by incorporating externally retrieved information as reliable clues, which helps to reduce factual errors in knowledge-intensive tasks \cite{guu2020retrieval,lewis2020retrieval}. 
REALM \cite{borgeaud2022improving} augmented language model pre-training with a latent knowledge retriever, and decided what knowledge to retrieve and use during inference. 
Atlas \cite{izacard2022few} explored end-to-end training of both the retriever and LM, demonstrating superior performance in few-shot learning. Recent approaches focus on integrating LLMs with RAG to imitate human behaviors in web search, such as WebGPT \cite{nakano2021webgpt}, WebCPM \cite{qin2023webcpm} and WebGLM \cite{liu2023webglm}. Asia et al. \cite{asai2023self} further proposed a self-reflective RAG framework, which retrieved passages on-demand for response generation. 
Whereas, these approaches mainly focus on the singe-round QA that merely involves the current question. How to perform effective RAG in conversational settings that questions are dependent on the context is not well studied. 
In this paper, we propose conversation-level RAG (\texttt{ConvRAG}), which incorporates a conversational question refiner, coupled with a fine-grained retriever and a self-check based response generator for CQA. 

\vspace{-2mm}
\section{Dataset}
\vspace{-1mm}
\subsection{Dataset Construction}
\vspace{-1mm}
\textbf{Data Collection.} 
We adapt two well-known CQA datasets, namely QReCC \cite{anantha2021open} and TOPIOCQA \cite{adlakha2022topiocqa}, which comprise of open-domain conversations with context-dependent question-answer pairs. Noting the insufficiency of Chinese CQA datasets and limited topics, we explore a potential method to automatically construct a Chinese CQA dataset with a wide range of topics, types, and complexities. Specifically, we first meticulously sample 1229 question-answer pairs from the COIG \cite{zhang2023chinese} and Zhihu-KOL\footnote{\url{https://github.com/wangrui6/Zhihu-KOL}} datasets. COIG covers various exam subjects such as Chinese, English, Politics, Biology, History and Geology, while Zhihu-KOL involves open-domain question answering with diverse topics. Then, we construct intent transition chains inspired from previous studies \cite{ahmad2021unsupervised}, and incorporate the chain into instructions to guide GPT-4 \cite{achiam2023gpt} to transform the question-answer pairs into multi-round conversations. Finally, we sample some data for checking to ensure the quality.

\textbf{Data Annotation.} 
For existing datasets and our constructed dataset, we provide annotations for question reformulation, keyword extraction, and the helpfulness of the retrieved paragraph based on the conversational context and the current question. For question reformulation and keyword extraction, we prompt ChatGPT with a specific instruction for automatic annotations. To ensure the quality, a small set of annotations is sampled for manually checking, which indicates the effectiveness of ChatGPT in question reformulation and keyword extraction over CQA. As ChatGPT has a poor performance in labeling the helpfulness of the retrieved paragraph, three professionally trained workers are asked to independently evaluate each search result as `useful' or `not useful'. The label with the majority votes is the final assessment result.
\vspace{-2mm}
\subsection{Dataset Analysis}
 We divide the samples from existing datasets (QReCC and TOPIOCQA) into three distinct parts, namely training, validation and test set. Since the conversational topic distributions in the test set are the same as that in the training set, we mark it as the test (seen) set, indicating that the topics have been seen during training. To better evaluate the generalization of our approach in CQA, we additionally utilize the constructed dataset with different topics and question types for testing, denoted as test (unseen) that the topics have never been seen during training. The overall data statistics are shown in Table \ref{tab:5}. We expand with several new features compared with existing CQA datasets, including reformulated question, extracted question keyword, retrieved paragraphs and their helpfulness. Each question is reformulated based on the context to make the questioner's intent more explicit. We further extract keywords from the reformulated question, which contains an average number of 3 keywords. For each reformulated question, the helpfulness of about 3 paragraphs is labeled for retrieval-augmented generation.



\begin{table}[!t]
\footnotesize
\caption{{\footnotesize Statistics of the dataset. Q$^\prime$ denotes the reformulated question.}}
\vspace{-3mm}
\label{tab:5}
\begin{tabular}{ l c c c c } 
 \hline
 & Train & Valid & Test (seen) & Test (unseen)\\ 
 \hline
\# Conv & 7118 & 1015 & 2033 & 1229 \\
\# Turns/Conv & 8.62 & 8.48 & 8.49 & 4.78 \\
\# Tokens/Turn & 30.05 & 30.05 & 30.73 & 82.66\\
\# Keywords/Q$^\prime$ & 2.82 & 2.72 & 2.79 & 3.20\\
\# Paragraphs/Q$^\prime$ & 2.63 & 2.57 & 2.62 & 2.70\\ 
\hline
 
\end{tabular}
\vspace{-4mm}
\end{table}

\vspace{-2mm}
\section{Our Approach}
\vspace{-1mm}
The overall architecture of our approach is shown in Figure \ref{figure1}, which consists of three components: (1) \textit{Conversational Question Refiner} focuses on better question understanding by question reformulation and keyword extraction based on the interdependent context; (2) \textit{Fine-Grained Retriever} aims to retrieve the most relevant information from Webs for augmenting response generation; (3) \textit{Self-Check based Response Generator} filters the unhelpful information with self-check for more accurate response generation.

Formally, let $((C_{i-1}, q_i), r_i) \in \{X, Y\} $ be a sample in the dataset, where $ C_{i-1} = \{q_1, r_1, \ldots, q_{i-1}, r_{i-1}\}$ denote the conversation context with $i-1$ turns, $q_i$ and $r_i$ is the current question and response. The goal is to predict the response $r_i$ given preceding context $C_{i-1}$ and the current question $q_i$.

\vspace{-2mm}
\subsection{Conversational Question Refiner}
 Since the question in CQA usually contains concepts or pronouns that are interconnected with preceding context (e.g., the `battle' refers to `Hunayn' in the first turn as shown in Figure \ref{figure1}), we need to refine the question to make the questioner's intent more explicit. Specifically, we explore two strategies for question refinement, namely Question Reformulation and Keyword Extraction. 




\textbf{Question Reformulation.} As shown in Figure \ref{figure1}, the original question is reformulated as `\textit{When did the Battle of
Hunayn happen?}', which clarifies the `\textit{battle}' according to previous context. To obtain the reformulation, we initialize a pre-trained LLM as $\mathcal{M_Q}$, and fine-tune it for new question generation with a standard conditional language modeling objective that maximizes the likelihood: 
\vspace{-2mm}
\begin{equation}
\vspace{-1mm}
\small
\max_{\mathcal{M_Q}} \sum_{ (C_{i-1}, q_i) \in X } \log P_{\mathcal{M_Q}}({q'_i}|C_{i-1}, q_i),
\end{equation}
where $q'_i$ represents the reformulated question.

\textbf{Keyword Extraction.} To concentrate on the core question intent, we further extract keywords based on the question, the reformulation and context, which facilitates retriever to return the most relevant information. We also train an LLM based model $\mathcal{M_K}$ for keyword extraction:
\vspace{-2mm}
\begin{equation}
\small
\max_{\mathcal{M_K}} \sum_{(C_{i-1}, q_i) \in X } \log P_{\mathcal{M_K}}({K_i}|C_{i-1}, q_i, q'_i),
\end{equation}
where $K_i $ is the set of extracted keywords.


\vspace{-1mm}
\subsection{Fine-Grained Retriever}
To alleviate the hallucination problem of LLMs, we retrieve relevant information from Webs, which serves as evidence to support response generation. In particular, the retriever includes three stages.

\textbf{Document-Level Retrieval.} In the first stage, we leverage the keywords $K_i$ as the query to search relevant web documents/pages with a search engine (e.g., Google), resulting in a set of documents as $\{D_1, \ldots, D_M \}$.



\textbf{Paragraph-Level Recall.} In the second stage, each web document is decomposed into paragraphs as $D = \{p_1, \ldots, p_{|D|}\}$, and we aim to find the relevant paragraph according to the relevance with the reformulated question and keywords. For a paragraph $p_j$, the relevance score with the question is formulated as:
\begin{equation}
\small
s_{ij} = f(E(q'_i), E(p_j)) + \sum_{k \in K_i} f(E(k), E(p_j)),
\end{equation}
where $E(\cdot)$ is an embedding function based on FastText \cite{joulin2017bag}, $f(\cdot)$ is a score function such as cosine similarity.


\textbf{Paragraph-Level Rerank.} In the third phase, we further select the most relevant paragraphs by a reranker, which is a model integrated into BGE\footnote{\url{https://github.com/FlagOpen/FlagEmbedding}}. Unlike conventional embedding models, the reranker uses the reformulated question and paragraph as input and directly outputs the relevance score instead of embeddings, which shows great potential in retrieval. Finally, the top $n$ most relevant paragraphs are selected for the generator, denoted as $P_{top}$.


\vspace{-1mm}
\subsection{Self-Check based Response Generator}


Previous studies indicate that directly utilizing the retrieved results can mislead the model and adversely impact the performance \cite{ren2023investigating}. To address this issue, we present a self-check mechanism, which urges LLMs to rethink the helpfulness of the retrieved information and filter the irrelevant noise for response generation. As shown in Figure \ref{figure1}, the generator first checks the helpfulness of each paragraph, and then generates the response according to the helpful information and the encapsulated intrinsic knowledge. Specifically, we train an LLM based model $\mathcal{M_R}$ for helpfulness checking and response generation:
\begin{equation}
\small
\max_{\mathcal{M_R}} \sum_{ (C_{i-1}, q_i) \in X } \log P_{\mathcal{M_R}}(h_i, r_i |C_{i-1}, q_i, P_{top}),
\end{equation}
where $h_i$ represents the helpfulness description for $P_{top}$, and $r_i$ represents the response to the question.


\section{Experiments}

\subsection{Experimental Setup}
\textbf{Evaluation Metrics.} 
We employ different automatic metrics, including BLEU \cite{papineni2002bleu}, ROUGE \cite{lin2004rouge} and METEOR \cite{banerjee2005meteor} for response evaluation. Moreover, the GPT-4 model is used for pairwise comparison, which has exhibited good capability in mimicking human preferences in content evaluation by previous studies \cite{naismith2023automated}.


\noindent \textbf{Baselines.} 
We compare with the recent advanced baselines:
1) ChatGPT \cite{ouyang2022training} is prompted with one-shot example and used as a basic baseline for CQA without retrieval augmentation;
2) WebCPM \cite{qin2023webcpm} mainly focuses on single-round QA, and integrates interactive web search for RAG;
3) WebCPM$^*$ utilizes our reformulated question as the input;
4) Perplexity.ai\footnote{\url{https://www.perplexity.ai}}is a production search system that supports CQA, and generates responses based on the conversation history and retrieved results.


\noindent \textbf{Implementation details.} We use the Baichuan 2-7B \cite{yang2023baichuan} language model for the modules of question reformulation, keyword extraction and self-check based response generator. The batch size is set to 16. The learning rate is set to 2e-4 and we use the cosine learning rate decay with 200 warm-up steps. After paragraph-level rerank, we obtain the top 3 relevant paragraphs as $P_{top}$.

\begin{table}[!t]
\vspace{-2mm}
\footnotesize
\scriptsize
\caption{The results of automatic evaluation. $*$ means the model utilizes our reformulated question as the input.}
\label{tab:main_results}
\vspace{-2mm}
\setlength{\tabcolsep}{3.2pt} 
\begin{tabular}{@{}clcccccccccc@{}}
\toprule
\multirow{2}{*}{Data} & \multirow{2}{*}{Model} & \multicolumn{4}{c}{BLEU} & \multirow{2}{*}{METEOR} & \multicolumn{3}{c}{ROUGE}\\ 
\cmidrule(lr){3-6} \cmidrule(lr){8-10}
                      &                        & 1     & 2     & 3     & 4     &                         & 1     & 2     & L                      \\ 
\midrule
\multirow{5}{*}{\makecell{Test \\ (seen)}} & ChatGPT                & 10.77 & 7.57  & 5.97  & 4.93  & 27.60                  & 26.51 & 10.91  & 17.87   \\
                      & WebCPM                 & 12.32 & 5.85  & 3.56  & 2.39  & 12.91                   & 18.55 & 3.94 & 14.17                  \\
                      & WebCPM$^*$       & 13.18 & 7.04  & 4.43  & 2.95  & 16.75                   & 20.87 & 5.61 & 15.31                 \\
                      & Perplexity.ai        & 26.89 & 20.34 & 17.03 & 14.78 & 23.37                   & 29.14 & 14.57 & 22.40                 \\
                      & \texttt{ConvRAG}                   & \textbf{32.96} & \textbf{25.34} & \textbf{21.33} & \textbf{18.58} & \textbf{30.75}                   & \textbf{40.53} & \textbf{21.94} & \textbf{32.40} &            \\
\midrule
\multirow{5}{*}{\makecell{Test \\ (unseen)}} & ChatGPT              & 11.79 & 8.33  & 6.22  & 4.81  & \textbf{33.36}                  & 35.12 & 12.40 & 18.42                  \\
                        & WebCPM               & 21.40 & 13.56 & 9.86  & 7.56  & 29.47                   & 34.53 & 12.20 & 23.62     \\
                        & WebCPM$^*$     & 22.33 & 14.49 & 10.60 & 8.12  & 32.94                   & 37.54 & 13.79 & 25.20                       \\
                        & Perplexity.ai      & 32.82 & 24.07 & \textbf{19.05} & \textbf{15.46} & 32.51                   & 44.02 & 20.92 & 30.66         \\
                        & \texttt{ConvRAG}                 & \textbf{33.90} & \textbf{24.15} & 18.67 & 14.90 & 33.23                   & \textbf{47.69} & \textbf{21.43} & \textbf{33.59}  \\
\bottomrule
\end{tabular}
\vspace{-4mm}
\end{table}

\vspace{-2mm}
\subsection{Main Results}
\textbf{Automatic Evaluation.} 
Table \ref{tab:main_results} shows the experimental results, and we have the following observations. \textbf{First}, our \texttt{ConvRAG} achieves overwhelming superiority across all evaluation metrics on the test (seen) dataset. In particular, the maximum improvement is up to 10\% regarding to ROUGE-L.
\textbf{Second}, \texttt{ConvRAG} maintains a leading position in most cases on the test (unseen) dataset, indicating the good generalization of our approach for open-domain CQA with new topics.
\textbf{Third}, though the baseline WebCPM merely supports single-round question answering, its performance can be boosted with our reformulated question (WebCPM$^*$).
All these findings verify the effectiveness of our approach in CQA and showcase the great potential compared with existing RAG methods.

\textbf{Pairwise Comparison.}
In addition to the traditional metrics used in text generation, the GPT-4 model has exhibited good capability in mimicking human preferences in content evaluation \cite{naismith2023automated}. Therefore, we adopt it to further ascertain the superiority of the responses generated by our approach and the best baseline Perplexity.ai. 
The win counts by GPT-4 evaluation are shown in Table \ref{tab:chatgpt-score}. We observe that our \texttt{ConvRAG} is superior to the baseline Perplexity.ai in both datasets. 
Given that our \texttt{ConvRAG} is based on the Baichuan 2-7B, while Perplexity.ai utilizes the InstructGPT with larger-scale parameters, these findings indicate the effectiveness of our fine-grained retrieval-augmentation and self-check to boost LLMs for CQA.


\begin{table}[!t]
\vspace{-2mm}
\footnotesize
\centering
\caption{Win Counts: \texttt{ConvRAG} vs. Perplexity.ai .}
\label{tab:chatgpt-score}
\vspace{-2mm}
\setlength{\tabcolsep}{4.0pt}
\begin{tabular}{lcc}
\toprule
\rowcolor{blue!5}
\textbf{Model} & Test (seen) & Test (unseen) \\
\midrule
\rowcolor{blue!20}
\texttt{ConvRAG} & 1143 & 656 \\
\rowcolor{blue!10}
Perplexity.ai & 890 & 573 \\
\bottomrule
\end{tabular}
\vspace{-2mm}
\end{table}

\begin{table}[!t]
\footnotesize
\caption{Ablation studies on Test (seen).}
\label{tab:Ablation}
\vspace{-2mm}
\setlength{\tabcolsep}{3.2pt} 
\begin{tabular}{lcccccccccc}
\toprule
\multirow{2}{*}{Methods} & \multicolumn{4}{c}{BLEU} & \multirow{2}{*}{METEOR} & \multicolumn{3}{c}{ROUGE} \\
\cmidrule(lr){2-5} \cmidrule(lr){7-9} \cmidrule(lr){10-11}
        & 1       & 2       & 3       & 4       &          & 1       & 2      & L     \\
\midrule
\texttt{ConvRAG} & \textbf{32.96} & \textbf{25.34} & \textbf{21.33} & \textbf{18.58} & \textbf{30.75} & \textbf{40.53} & \textbf{21.94} & \textbf{32.40}\\
\textit{- QF} & 27.98 & 21.18 & 17.68 & 15.30 & 28.92 & 38.44 & 20.00 & 30.25 \\
\textit{- FR} & 27.17 & 20.78 & 17.42 & 15.11 & 29.91 & 39.56 & 21.20 & 31.23 \\
\textit{- SC} & 32.14 & 24.45 & 20.75 & 17.78 & 29.50 & 39.28 & 20.55 & 31.36 \\
\textit{- ALL} & 27.63 & 20.78 & 17.26 & 14.91 & 29.27 & 39.22 & 20.93 & 30.93 \\
\bottomrule
\end{tabular}
\vspace{-2mm}
\end{table}

\subsection{Ablation Studies}
We conduct ablation studies to investigate the effectiveness of each component of our approach, and the results on Test (seen) are shown in Table \ref{tab:Ablation}. Similar results can be observed on Test (unseen), and we do not present them for the limited space. The notations \textit{- QF}, \textit{- FR} and \textit{- SC} mean removing the component as Conversational Question Refiner, Fine-Grained Retriever and Self-Check in the response generator respectively. For \textit{- FR}, we utilize the retrieved snippet instead following previous studies \cite{asai2023self}. 

We observe that the absence of any component leads to a decline in the performance, indicating the effectiveness of all components coupled with LLMs to help boost CQA. In particular, the conversational question refiner (\textit{QF}) plays a more important role, which results in about 2 points in terms of METEOR and ROUGE-L. This corresponds to previous findings that it is necessary to perform question rewriting or reformulation in CQA \cite{anantha2021open}, since users usually use certain concepts or pronouns that are interconnected with preceding context, which poses a big challenge to understand the question in CQA. Thus, we present the conversational question refiner, which performs question reformulation and keyword extraction for better question understanding. The second important components are fine-grained retriever and self-check, which further filters irrelevant information or noise for response generation.

\section{Conclusions and Further Work}
In this work, we propose \texttt{ConvRAG} to enhance CQA with fine-grained retrieval-augmentation and self-check. 
Extensive experiments on both the benchmark and our constructed dataset demonstrate the superiority of our approach over the advanced baselines. 
The ablation studies also verify the effectiveness of each component of \texttt{ConvRAG}, especially for the conversational question refiner.
Moreover, we construct a Chinese CQA dataset extending with new features and diverse topics for RAG research. 
In the future, we will explore how to augment LLMs with knowledge bases and adapt our approach for more conversational settings.

\newpage
\clearpage
\bibliographystyle{ACM-Reference-Format}
\bibliography{sample-base}

\end{document}